\pdfoutput=1
\documentclass[11pt]{article}

\usepackage[]{acl}

\usepackage{times}
\usepackage{latexsym}
\usepackage{graphicx}

\usepackage[T1]{fontenc}
\usepackage[utf8]{inputenc}

\usepackage{microtype}

\newcommand{\nop}[1]{}

\title{Developing a Production System for Purpose of Call Detection in Business Phone Conversations}

\author{Elena Khasanova, Pooja Hiranandani, Shayna Gardiner, \\
{\bf Cheng Chen, Xue-Yong Fu, Simon Corston-Oliver} \\
  Dialpad Canada Inc \\
  1100 Melville St \#400 \\
  Vancouver, BC, Canada, V6E 4A6 \\
  \texttt{ \{elena.khasanova,phiranandani,sgardiner\}@dialpad.com} \\
   \texttt{\{cchen,xue-yong,scorston-oliver\}@dialpad.com} \\}

\begin{document}
\maketitle
\begin{abstract}
For agents at a contact centre receiving calls, the most important piece of information is the reason for a given call.  An agent cannot provide support on a call if they do not know why a customer is calling.  In this paper we describe our implementation of a commercial system to detect \textit{Purpose of Call} statements in English business call transcripts in real time. We present a detailed analysis of types of Purpose of Call statements and language patterns related to them, discuss an approach to collect rich training data by bootstrapping from a set of rules to a neural model, and describe a hybrid model which consists of a transformer-based classifier and a set of rules by leveraging insights from the analysis of call transcripts. The model achieved 88.6 F1 on average in various types of business calls when tested on real life data and has low inference time. We reflect on the challenges and design decisions when developing and deploying the system. 

\end{abstract}

\section{Introduction} \label{intro}

The Purpose of Call as we define it is similar to a thesis statement in an argument: it introduces the speaker’s intent, the broad theme or topic of a conversation, any key entities, and relevant relationships between them. The Purpose of Call statement might also include a linguistic signpost -- an indication to the listener that the utterance is intended to convey the purpose of the speaker’s call.

For instance: 

I’m \textit{calling because}  [signpost] I'm trying to open \textit{one of the programs} [entity] on \textit{my computer} [entity] and \textit{it's not opening} [relation] so I’m hoping I can \textit{get some assistance} [intent] with that. \footnote{In contrast, statements such as \textit{I’m calling to ask a question} are not considered Purpose of Call expressions even though they contain relevant signposting language because there are no entities an agent or a customer can relate to.}
 
Purpose of Call statements in a contact centre setting are usually uttered by the customer in inbound calls, and by the agent in outbound calls. The Purpose of Call is typically stated near the beginning of the call, is often stated in a single utterance, and does not contain extra information.  Atypically, we may sometimes see the Purpose of Call occurring in the middle of a conversation, occurring across several utterances, being implicit, or being uttered by a call recipient rather than a call initiator. 

\begin{figure}[t]
\centering
\includegraphics[width=\linewidth]{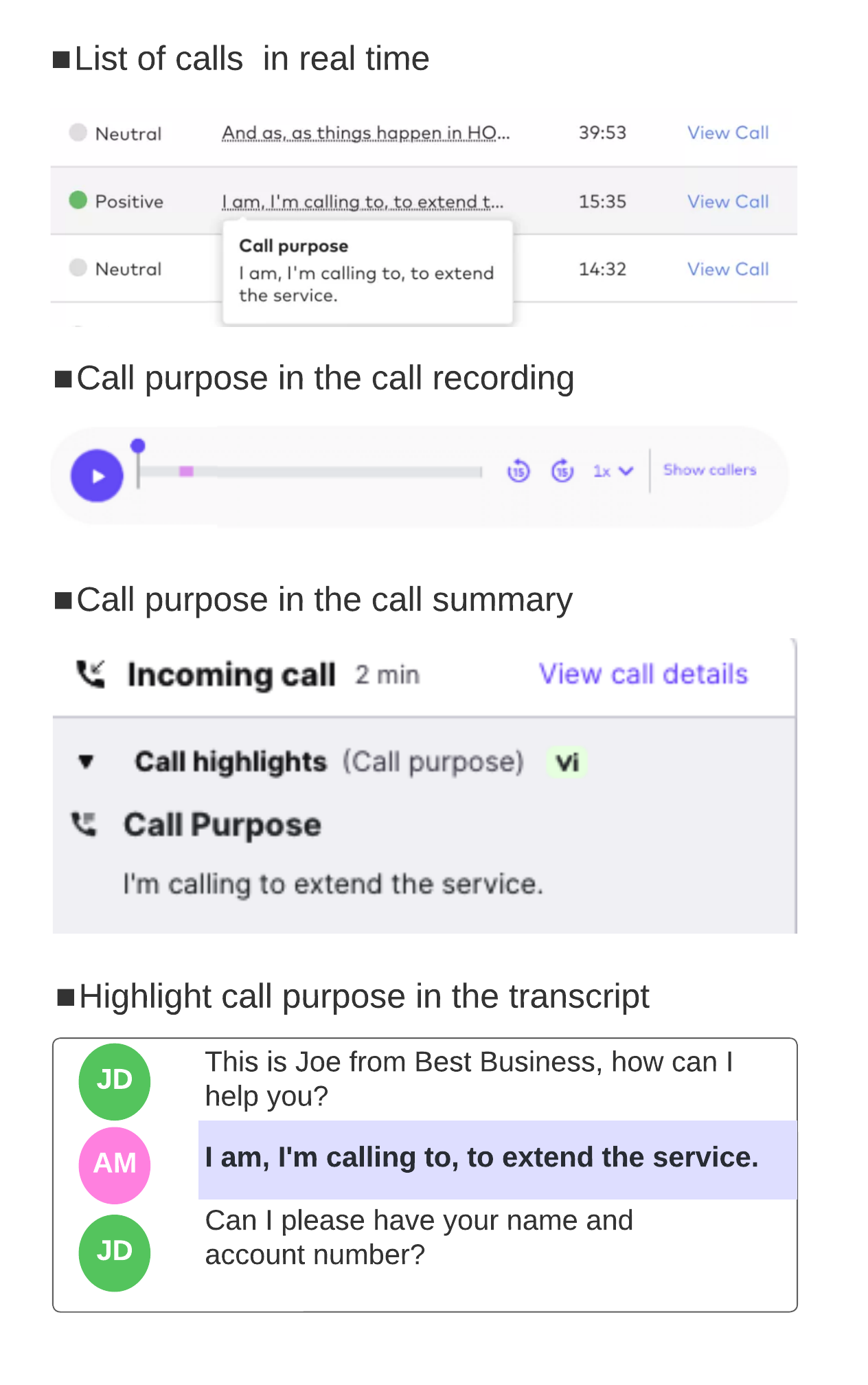}
\caption{An illustration of the applications of Purpose of Call}
\label{fig1}
\end{figure}

The models described below have been implemented in the Dialpad Contact Center product and are running in production. The Purpose of Call is extracted from an automatic speech recognition (ASR) generated transcript in near-realtime and displayed in a dashboard used by call center supervisors to monitor calls taking place. The dashboard shows information about the caller and the agent, the duration of the call, the Purpose of Call, and customer sentiment. A separate analytics component clusters the Purpose of Call from all calls in a call center during a time period to provide insights about trends and anomalies, customer pain points, and common problems and knowledge gaps among agents. Additional use-cases include showing the Purpose of Call in a summary of prior calls with a customer, and including the Purpose of Call in summaries of the conversation. The utterance segment containing the Purpose of Call is highlighted in the call transcript and the call recording to be easily accessible to agents and call center supervisors. These use-cases are summarized in Figure \ref{fig1}. The Purpose of Call feature is used to help call center managers to navigate to relevant sections of conversations to identify areas to coach sales and support agents and sample relevant calls. Through customer education, we emphasize that the feature should not be used for automated evaluation of agent performance.

There are a few challenges that arise when building an automatic system to detect Purpose of Call. 

\textbf{Diversity of Purpose of Call statements.} This type of detection system should be an open-class system, since call purposes vary across different domains, industries, and types of calls.

\textbf{Robustness to noise.} There are challenges related to the fact that Purpose of Call extraction relies on the output of an ASR system. There are two sources of noise in the ASR transcripts: language production issues such as false starts, dysfluencies, filled pauses, inconsistency in conversational turn-taking ~\cite{Cailliau2013MiningAS, dutrey:hal-01134812, zelasko2019better, Clavel2013SpontaneousSA} as well as their representation in the ASR system and recognition errors due to acoustic noise.

\textbf{Limitations in training data.} Existing intent detection datasets do not reflect real world settings (e.g. they do not distinguish the Purpose of Call from other intent-like statements, and are limited to a subset of domains). Manually annotating data (e.g. using crowd-sourced annotators) raises privacy concerns since annotators must have access to a full conversation transcript in order to find the best Purpose of Call. Annotation is a complex task that requires highly trained annotators, and is expensive and  time-consuming because annotators must consider the larger context of the conversation to make a judgment.

\textbf{Computational efficiency.} The need for the system to extract the Purpose of Call statement in real time imposes constraints  on memory consumption, latency, and inference speed.

This paper describes an end-to-end system to extract a Purpose of Call statement from the transcript of a business telephone call. Our contributions are three-fold:

\textbf{1. Data analysis}: we present a detailed analysis of language patterns and other features involved in call purpose detection;

\textbf{2. Data}: we describe a process to overcome a lack of training data by bootstrapping a deep learning model from a knowledge-engineered model and discuss the heuristics for developing such a model;

\textbf{3. System}: we describe optimizations that were done in an online commercial system to identify Purpose of Call statements in near-realtime (within three seconds of an utterance being transcribed). To evaluate the effectiveness of our approach, we examine the actual output of our production system. 

% \textcolor{blue}{It should be noted that comparison with other baseline systems in terms of performance or computational efficiency is not in the scope of this paper; the Purpose of Call detection task is different from potential candidates for comparison, such as intent detection or dialogue act modeling systems. An important distinction is  the possibility of multiple utterances conveying various intents in a conversation and only one utterance with specific linguistic characteristics serving as a Purpose of Call statement representative of the conversation. Dialogue act modeling systems also depend on the tag sets they utilize (e.g. \cite{10.1145/3025171.3025191}), which are different from our approach grounded in Conversation Analysis framework. To make this comparison appropriate, the baseline models would need to be adapted to the domain, the data, and the task, which would essentially require building new models.} 

\section{Related work} \label{literature}

The concept of a Purpose of Call statement has its roots in the Conversation Analysis framework~\cite{Schegloff1973,Sacks1974}.  Within this framework, which combines perspectives from Linguistics and Sociology, a conversation is understood to be composed of turn-taking utterances, with each “turn” being indicated via linguistic and paralinguistic cues. Conversational turns often form adjacency pairs such as question-answer pairs or offer-acceptance/refusal pairs. There are other key aspects of a conversation as well. For instance, a conversation is likely to end after a linguistic cue known as a "closing" is given; likewise, there is usually a linguistic indicator that a conversation is being initiated: an opening ~\cite{Schegloff1973,Sacks1974,Pomerantz2011}. Most work analyzing telephone conversation openings within the framework of Conversation Analysis has been conducted on English, but similar patterns have been observed in other languages, including German and Farsi \cite{TaleghaniNikazm2002ACA}.

Within a contact center environment, the Purpose of Call is, like openings and closings, an integral aspect of the conversation (i.e. call) between customer and support agent. We propose that a Purpose of Call is a particular conversational feature that is necessary in contact center calls, and is distinct from the call opening, the body of the call, and the call closing.

The first 120 seconds of a customer support call are predictive of that call’s outcome ~\cite{takeuchi-etal-2007-automatic, article}. The Purpose of Call statement typically occurs within this timeframe, so highlighting a Purpose of Call in real time could provide agents with additional support in meeting customer needs.

% \begin{table*}
% \centering
% \begin{tabular}{p{0.35\linewidth} | p{0.6\linewidth}}
% \hline
% \textbf{Context} & \textbf{Use case} \\
% \hline
% Call center agent & log call purpose details into a ticket tracking system, use call purpose information from a previous call to prepare for subsequent interactions, review similar call purposes and consolidate related problems with their appropriate solutions\\
% \hline
% Call center manager & realtime monitoring of calls, more effective coaching, assess common problems, identify knowledge gaps, evaluate where agents are following a certain playbook or communication protocol as it relates to specific call purposes, assess trends and anomalies, customer pain points, needs of the call center\\
% \hline
% Downstream systems & conversation chunking, conversation summarization, highlighting different conversational aspects of a call\\
% \hline
% \end{tabular}
% \caption{Example use cases of the call purpose detection feature }
% \end{table*}

\section{Methodology} \label{method}
We formulate the Purpose of Call detection task as a binary classification problem. Each call, after being transcribed by the ASR system, is represented as a sequence of utterances, which may consist of one or more sentences. The division into utterances is based on acoustic features such as silent pauses and the length of a speech fragment.

For a given utterance, we determine the probability that the utterance is the Purpose of Call statement for that particular call. We impose the following constraints on this task: (i) For a given call, there is only one most probable Purpose of Call. (ii) Only calls with two call sides (agent and customer) are considered, which excludes multiparty business conversations. (iii) The model should make a prediction as the call is ongoing and therefore will not have access to the full conversation.
% such as company meetings conducted using [REDACTED] telephony products. 

Due to the lack of available annotated data representing the concept of Purpose of Call, we followed an iterative approach to develop the model, consisting of three steps:
(1) Computational Linguists on-staff conducted extensive linguistic analysis of transcripts to identify the characteristics of Purpose of Call statements. (2) We then implemented a knowledge-engineered approach to identify these Purpose of Call statements. (3) We bootstrapped from the knowledge-engineered solution, using it to label training data for a transformer-based approach. We select a transformer-based model as it is the current state-of-the-art in sequence classification and is known to have better generalization power than rule-based models.

We evaluate the performance using F1, Precision, and Hit rate, i.e. the number of calls in which a Purpose of Call was detected out of all available calls. We measure Hit rate in calls at least 30 seconds long, based on the observation that shorter calls may not include any content (e.g. because the caller hung up before starting the conversation). The model is tested on an automatically obtained validation set that represents 10\% (18K utterances) of the training data, a manually annotated gold test set of 13215 utterances from 909 calls, and unlabeled samples from 600 real-life calls. 

\subsection{System Overview} \label{system}

\begin{figure*}[t]
\centering
\includegraphics[width=\linewidth]{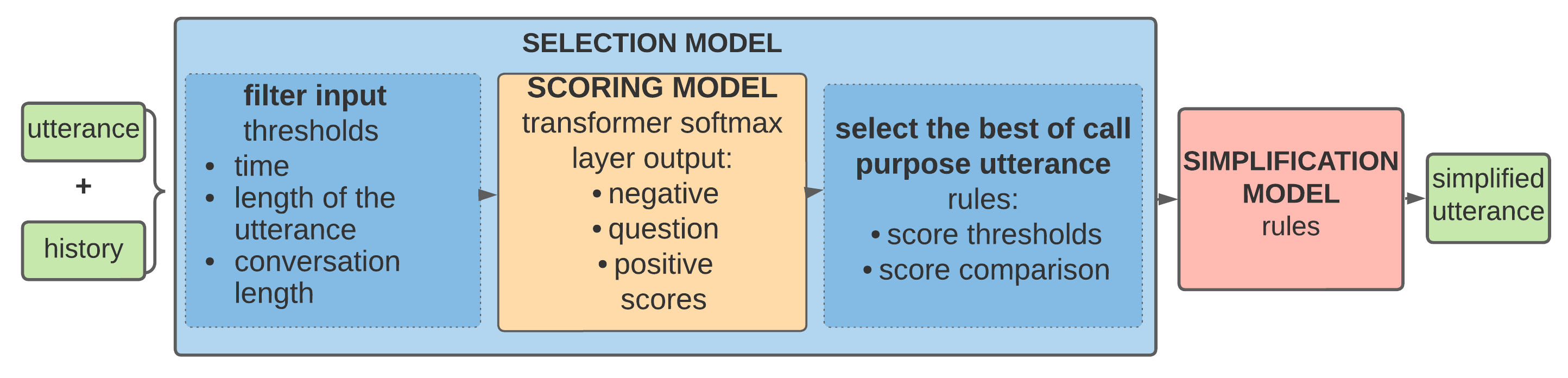}
\caption{Purpose of Call Detection System Architecture}
\label{fig2}
\end{figure*}

The production system to detect a Purpose of Call utterance is a hybrid model consisting of three parts (see Figure~\ref{fig2}).

\textbf{The Selection model}, or outer model, inputs an utterance, the previous context of the conversation, and the probabilities of previously detected Purpose of Call events. It consists of two sets of rules: (1) empirically derived filters that determine whether an incoming utterance is a candidate for a Purpose of Call and should be processed by the inner model, a successful candidate is within 180 seconds and 30 utterances in the call, and is between 4 and 150 tokens long; (2) rules that combine and compare scores from the inner model and set various thresholds for different linguistic types of call purpose statements (i.e. the utterances containing signposting language typically receive higher scores than other types and need a higher bar). Every time a new utterance qualifies to be a Purpose of Call, it is dynamically updated in the user interface. (See Appendix ~\ref{sec:appendix_selection} for an example of a Selection rule.)

\textbf{The Scoring model}, or inner model, is implemented as a multiclass classification model which performs inference on a single utterance. We fine-tuned a transformer-based model for classification on proprietary labeled data. The model assigns to an utterance probabilities of it being a \textit{call purpose},  \textit{question}, or \textit{negative} (not a \textit{call purpose} or a \textit{question}). The \textit{question} class represents \textit{question\_response} pattern (see Table \ref{cp_types}) and is used to boost probabilities of utterances that would otherwise be of the \textit{negative} class.

\textbf{The Simplification model}. The utterance with the highest score is stripped of information that is irrelevant to the purpose of the call (e.g. \textit{greetings, pleasantries, introductions, technical problems}). It consists of a small set of common expressions (many of which are reused from the knowledge-engineered model) to exclude from utterances and reduces the length of Purpose of Call utterances by 7.8\% on average. 49\% of utterances undergo simplification with precision of 96\%. 

% See an example (text in red is being removed by the model): 

% \textit{{\color{red}Doing great, thank you. Good afternoon. This is Benjamin with Company-XYZ. Yes, I can hear you.} I’m reaching out about our new product offer. {\color{red}Is this a good time to call?}}

\subsection{Model Development} \label{model_dev}
The model development process consists of three main stages: data analysis and feature selection, designing a knowledge-engineered model informed by the insights from the data, and bootstrapping a transformer based model from the rule-based system. In this section, we discuss these stages.
\subsubsection{Data Research and Feature Selection} \label{data_an}

We manually analyzed a sample of 2000 call transcripts across several dimensions, which are outlined below.

\begin{enumerate}
    \item \textbf{Inbound vs. outbound calls}:\\
    In \textit{outbound} call center calls (40.8\%), the call initiator and the side that utters the Purpose of Call is usually an agent; in \textit{inbound} calls (59.2\%), it is a customer, with some exceptions. 55.3\% of all Purpose of Call statements are uttered by the customer.
    \item \textbf{Place}: the Purpose of Call is uttered within the first 40 seconds in a majority of calls (73.8\%), in the middle in a minority of calls (10.7\%), and towards the end in a handful of (mainly short) calls. The \textit{mean} time of occurrence is 29.9 seconds, \textit{std}=19.1,  \textit{median}=25.5 seconds. The \textit{maximum} time is 180 seconds.
    \item \textbf{Speaker role}:\\
The \textit{call initiator} utters the Purpose of Call in the vast majority of cases in the form of a \textit{statement}; in returned or scheduled calls, the Purpose of Call can be uttered by a \textit{call recipient} in the form of a \textit{guess, assumption or inquiry}.

\item \textbf{Domain}:
There are three main types of call center calls:\

\textbf{Sales calls}: commonly characterized by the Purpose of Call \textit{not} being stated explicitly in one utterance, but gradually being revealed during the course of the call. Agents often spend a longer time building rapport, so utter the purpose later in the call compared to support calls. 74\% of Purpose of Call statements still occur within the first 40 seconds. Outbound calls are prevalent.  48\% of Purpose of Call statements are uttered by customers.
\\
\textbf{Support calls}: inbound calls are prevalent, the Purpose of Call is introduced early in the conversation. In fact, 56\% of Purpose of Call statements are in the very first utterance, accompanied by a \textit{greeting}, and 63\% of the statements occur within the first 50 seconds. 62\% of Purpose of Call statements are uttered by customers.
\\
\textbf{General business calls}: may include support and sales calls as well as other communications, both formal and informal. The Purpose of Call is often implied (e.g. a conversation between colleagues, transfers from a chat to a call, with the purpose of the conversation being known by both parties). 84\% occur within the first 45 seconds, and 59\% are uttered by customers.

\item \textbf{Length distribution}:  Purpose of Call utterances range in length from 4 to 224 tokens, with the \textit{mean}=45.5, \textit{std}=29.9, \textit{median}=37, and 75\% being under 59 tokens. 
\item \textbf{Language patterns}:
We identified several language markers associated with Purpose of Call statements in Table~\ref{cp_types}.

\begin{table*}
\centering
\begin{tabular}{p{0.2\linewidth} | p{0.03\linewidth} | p{0.35\linewidth} | p{0.3\linewidth}}
\hline
\textbf{Pattern} & \textbf{\%} & \textbf{Description} & \textbf{Example}\\
\hline
call\_purpose\_phrases & 32.7 & explicit declarations of the Purpose of Call typically signposted with lexical cues containing \textit{purpose} and \textit{call} and their synonyms & \textit{The \textbf{reason for my call} is I moved to a new address, so I need to change it on my profile.}\\
\hline
desire\_phrases & 31.7 & expressions of volition, desire or need & \textit{Hi, I \textbf{need} a refund for my order. }\\
\hline
question\_response & 15.8 & responses to an agent’s prompt & \textit{- \textbf{How can I help you?} / - I received a message that my order has been delayed.}\\
\hline
greetings & 9.1 & long statements of at least 30 word tokens that follow a \textit{greeting} and occur within the first 6 utterances in the conversation &
\textit{\textit{Hey}, this is Christine. There is a police report, it was next to you guys why you heard it <...>}\\
\hline
problem\_phrases & 4.4 & express problems and concerns & \textit{I’m \textit{having an issue} with the delivery.}\\
\hline
update & 5.8 & updates and announcements & \textit{I \textbf{have an update} on your passport status.}\\
\hline
continuation & 0.4 & questions preceded by a signpost in the same utterance or a subsequent one from the same speaker & \textit{Hi, I'm calling because I \textbf{have a question}. / Do you accept new patients?}\\
\hline
\end{tabular}
\caption{\label{cp_types}
Language patterns in Purpose of Call statements
}
\end{table*}

\end{enumerate}

Approximately 7\% of calls in this sample do not contain an explicit Purpose of Call statement. Instead, the participants in the call appear to already have the context necessary to understand the call purpose.

\subsubsection{Knowledge-Engineered Model} \label{ke_model}

As outlined in Section~\ref{intro}, collecting labeled data for Purpose of Call extraction is a challenging task. Therefore, to obtain a representative sample of training data, we first implemented a knowledge-based model that takes into account the following parameters: \textit{utterance length} in tokens, the \textit{order} of an utterance in the conversation, the \textit{history} including several preceding utterances, and the presence or absence of \textit{language patterns} summarized in Table~\ref{cp_types} and implemented using regular expressions syntax (see  Appendix~\ref{sec:appendix_extract_poc} for an example). In total, stemming from the analysis in Section~\ref{data_an}, 8 rules (56 regex patterns) to detect call purpose candidates and 6 rules (55 regex patterns) to filter out negative statements were developed. The model reached a precision of 90.8\% and hit rate of 77\% on average across three domains (see Table ~\ref{eval}).

Further, we conducted error analysis by manually labelling the output of the production system on a random sample of 1000 calls. 
% For 23% of
% calls in the sample, no utterance was identified as the Purpose of Call. After human review, we determined that 13.5% of those calls did not contain an identifiable Purpose of Call and could be considered true negatives, and 37.5% of missed hits can be attributed to ASR errors. 27.6% of false negatives include cases with the Purpose of Call being known prior to the conversation (e.g. from shared knowledge, logged information, or in return calls)
After human review, we determined that 3\% of calls did not contain an identifiable Purpose of Call and could be considered true negatives, while 20\% were false negatives. 40\% of these false negatives can be attributed to ASR errors. 27.6\% of false negatives include cases with the Purpose of Call being known prior to the conversation (e.g. from shared knowledge, logged information, or in return calls) and therefore not considered by the model, 9.1\% correspond to specific industries (e.g. transportation) underrepresented in the data used in the analysis, and 44.7\% were caused by the limitations of the rules (note that these groups of false negatives intersect, hence the percentages do not add up to 100\%). False positives were mainly related to the lack of morphological flexibility in the rules and speech dysfluencies. In 6.2\% of calls, several utterances were legitimate Purpose of Call statements and the one selected by the model was not the best one.
These findings motivated the need for a transformer-based model that was more forgiving of ASR noise, had better generalization power, and was more responsive to changes in the data.

\subsubsection{Transformer-Based Model} \label{transformer}

\textbf{Training Data Collection.} \label{data_col}
Since the knowledge-engineered model achieved high precision, we could rely on its output to train a deep learning model. The dataset consists of English language utterances obtained from business calls in a variety of industries, with accompanying metadata such as timestamps for each token, call side, and call id. See Appendix~\ref{appendix_data} and \ref{data_an} for detailed statistics.
We randomly sampled one million utterances between February 6, 2020 and February 22, 2021, allowing only those that meet the requirements for a Purpose of Call candidate in \ref{system}. The utterances were divided into two sets: (1) those from the calls with a Purpose of Call hit (likely to be a true positive), (2) utterances from calls with no hit (may contain false negatives). With a series of patterns, we filtered out utterances that are likely to be false positives based on error analysis in \ref{ke_model}. Further, we sampled several datasets of 180K utterances with varying label and language pattern distributions in order to experimentally find the best configuration (see Appendix ~\ref{appendix_data}).
A train, development, and validation split of 80/10/10\% of data was used in each experiment. In addition, we created a golden dataset of 909 manually labeled calls, with the utterances organized chronologically within the call and limited to up to 30 utterances per call. This sample comprises 13215 utterances.

\begin{figure}[t]
\centering
\includegraphics[width=1.0\linewidth]{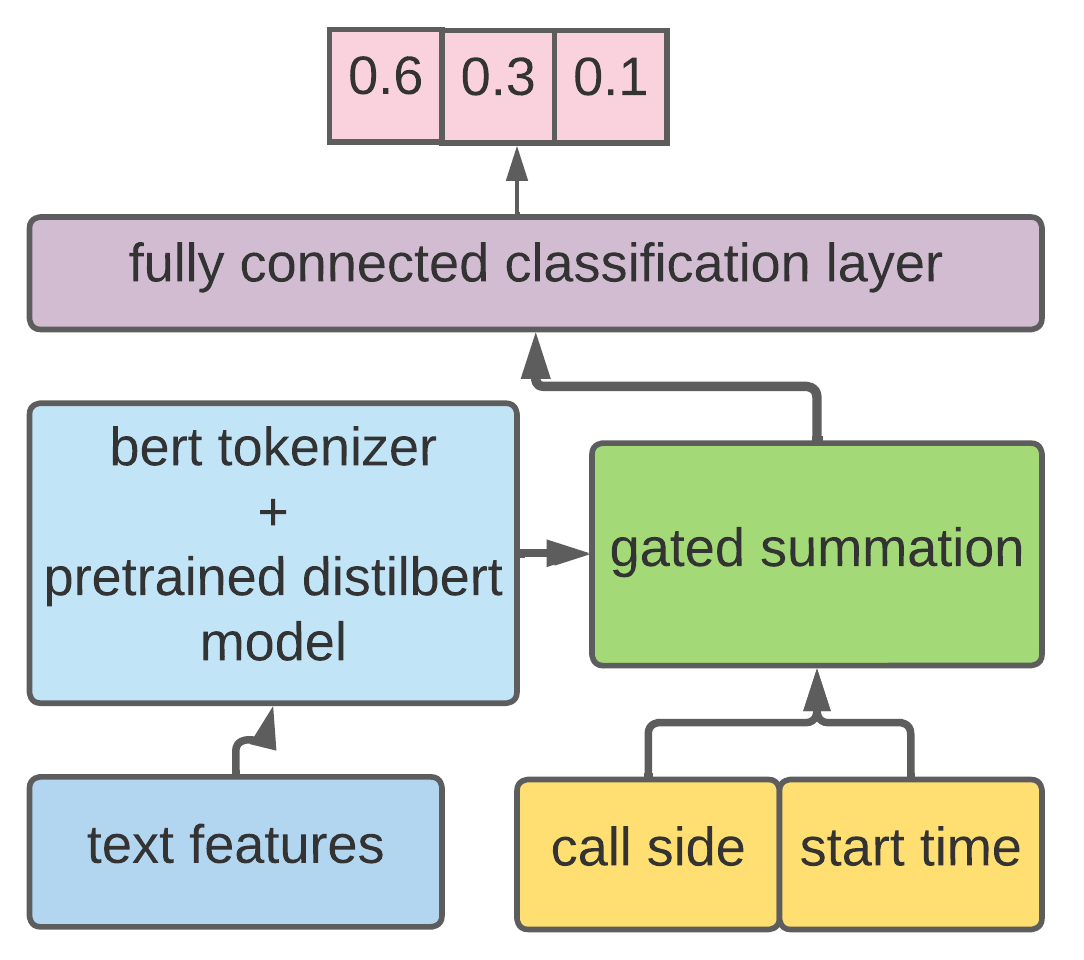}
\caption{Multimodal Transformer based scoring model}
\label{fig3}
\end{figure}

\textbf{Training Details.} We employ the DistilBERT~\cite{DBLP:journals/corr/abs-1910-01108} model, trained for classification with multimodal features. We combine text features with numerical and binary features, utterance \textit{start time} and \textit{call side} respectively, which have proven to be useful in the knowledge-engineered model, and pass on a gated summation of the transformer output with these features to the classification layer, following the approach in~\cite{gu-budhkar-2021-package}. This configuration outperformed other base models \footnote{https://huggingface.co/microsoft/DialoGPT-small} \footnote{https://huggingface.co/DeepPavlov/bert-base-cased-conversational}, combinations of multimodal features, and combining mechanisms outlined in~\cite{gu-budhkar-2021-package}. The model architecture is shown in Figure ~\ref{fig3}. We implement a data-driven iterative fine-tuning process with extensive error analysis and data resampling. See Appendix ~\ref{appendix_training} for details.

\subsection{Model Deployment}

Since the model was to operate in a near-realtime environment as a call is ongoing, optimising for inference time was a dominant consideration during model design. The model would perform inference on one CPU core. The model would need to accommodate the time taken to transcribe voice to text and properly format and punctuate the transcription, many of these tasks being accomplished by other deep learning models. 

Optimizations include: (i)  Having the Selection model that uses input features to filter utterances, thereby reducing the number of utterances that were attended to by the transformer model. These features include utterance count number and utterance start time, both of which should be below a threshold determined by experimenting with different parameter values in the knowledge-engineered model. (ii) Incorporating numerical and binary features into the deep learning model - adding signals beyond lexical features allowed us to use a lower capacity BERT variant with faster inference time. (iii) Capping input length to an empirically derived ceiling further reduced memory consumption and inference time.

The system was deployed in containers\footnote{https://cloud.google.com/kubernetes-engine} with 1 CPU and maximum 1GB memory per instance. The average inference time at the 95th percentile is 0.51 seconds, which meets the requirements of our production system for near-realtime deployments to complete inference in under 3 seconds.

\section{Evaluation} \label{eval}

Table \ref{performance} shows full results of the comparative evaluation of the knowledge-engineered and the hybrid models on business calls in three domains.  

\begin{table}
\centering
\begin{tabular}{lllll}
\hline
\textbf{Domain} & \textbf{Model} & \textbf{Precision} & \textbf{Hit Rate} & \textbf{F1}\\
Support & rules & 93.5 & 80.0 & 86.2 \\
 & hybrid & 91.0 & 90.4 & 90.7 \\
General & rules & 90.0 & 74.2 & 81.3 \\
 & hybrid & 89.0 & 85.6 & 87.3 \\
Sales & rules & 88.5 & 78.7 & 83.5 \\
 & hybrid & 87.0 & 88.9 & 87.9 \\
Avg & rules & 90.6 & 77.6 & 83.6 \\
 & hybrid & 89.6 & 88.3 & 88.6 \\
\hline
\end{tabular}
\caption{\label{performance}
Comparative evaluation of knowledge-engineered (here \textit{rules}) and hybrid models for Purpose of Call detection.
}
\end{table}

Qualitative analysis was conducted on the gold set: real life user data of 600 samples, and 200 calls with missed hits. False positives mainly correspond to signposting language without mention of the actual Purpose of Call, and indicate the model's over-reliance on lexical features. The model was found to be accurate in assigning utterances to classes, but not always sensitive to the difference between a \textit{valid} and \textit{the best} Purpose of Call. This can be addressed by introducing more contrastive examples in training data. Missed hits include cases initially excluded from the sample such as Purpose of Call stated across several utterances, and multiple Purpose of Call statements of equal importance. A synthesis of several utterances instead of selecting only one of them might be useful in such cases.

\section{Conclusion}

This paper discusses the development and deployment of a hybrid system to detect a Purpose of Call statement in business call transcripts for the English language in near-realtime settings. We introduce the concept of the Purpose of Call, provide in-depth analysis of real life data, and discuss overcoming the absence of available training data by bootstrapping from a knowledge-engineered model to a deep learning one. Both the knowledge-engineered and hybrid models demonstrate high precision and hit rate, with the hybrid model showing better performance while maintaining computational efficiency.

\section{Ethics Statement}

\textbf{Data}. The conversational data is presented in the form of individual utterances with sensitive data such as personal identifiable information removed. No crowdsourced annotation has been conducted, and access to the data was available only to a small number of in-house Scientists.

\textbf{Use}. The Purpose of Call feature is used by call center managers to identify areas to coach sales and support agents. It is recommended to not use this feature for automated evaluation of agent performance. Incorrect Purpose of Call prediction may provide unsatisfactory user experience for the managers as they sample calls but does not present any risk of negative impact for the agents.

\textbf{Licensing}. We follow the licensing requirements accordingly while using external tools such as HuggingFace \footnote{https://huggingface.co/} and Multimodal-Toolkit~\cite{gu-budhkar-2021-package} libraries.

\bibliography{custom}
\bibliographystyle{acl_natbib}

\appendix

\section{Appendix: Example Rule in a Knowledge-Engineered Model}
\label{sec:appendix_extract_poc}

An utterance is a Purpose of Call if:
\begin{itemize}
    \item It contains signposting phrases expressing a problem such as \textit{I'm having a problem, There is an issue, I'm having a hard time, I'm trying ... and it's not working}, 
    \item It occurs within the first 10 utterances
    \item It is at least 12 tokens long
  
\end{itemize}

Example: \textit{I got a really big problem here. When I log in, it asks for some pin, and I really, I can't use it. So there's obviously an issue here and can you help me with it?}.

\section{Appendix: Example Selection model heuristics}
\label{sec:appendix_selection}

% \item Pass an utterance to the scoring model if:
%     \begin{itemize}
%         \item the utterance ends within the first 180 seconds of the call
%         \item the absolute utterance count so far is <=30
%         \item the utterance is between 4 and 150 tokens long
%     \end{itemize}

Combine the \textit{positive} score for an utterance with the maximum \textit{question} score of the two preceding utterances in another call side. If it passes a threshold and is the biggest score so far, this utterance is a Purpose of Call.

\section{Appendix: Data Statistics}
\label{appendix_data}

\textbf{Total number of calls}: 86310 \\
\textbf{Total number of utterances}: 180 000 \\
\textbf{Industry distribution}: see Table ~\ref{ind}.

\begin{table}[hbt!]
\centering
\begin{tabular}{p{0.8\linewidth} | p{0.1\linewidth}}
\hline
\textbf{Industry} & \textbf{\%} \\
\hline
Technology & 25.1\\
IT, Consulting  & 15.5\\
Professional, Business Support Services & 14.1\\
Travel & 11.4\\
Health and Wellness & 5.6\\
Real Estate & 5.1\\
\hline
\end{tabular}
\caption{\label{ind}Industry distribution in training data: top 6 types}
\end{table}

\textbf{Label distribution:}
A key factor in training the model was determining the right distribution of labels and language patterns. The classes in our problem are naturally imbalanced: since only one utterance per call is a valid Purpose of Call, the vast majority of utterances are of the \textit{negative} class. In a random sample, only 4.7\% utterances are \textit{positive} hits, and only 1.9\% are \textit{questions}. If the data is sampled randomly, the model is likely to overfit to the \textit{negative} class. Sampling uniformly may reduce the number of complex instances in favor of the ones easier for the model to learn.
A set of experiments were conducted to determine the distribution of classes with the goal of optimizing accuracy of the Purpose of Call class predictions. We determined the optimal distribution of classes as follows: 42.5\% \textit{positive}, 42.5\% \textit{negative}, 15\% \textit{question} utterances (corresponds to the share of this pattern in real data). All utterances came from calls with a positive hit, which minimized the chance of false negatives in the training data.

\textbf{Language patterns distribution:}
From the error analysis and experiments, we determined the optimal distribution of language patterns within the \textit{positive} class:
\begin{itemize}
    \item 30\% \textit{call\_purpose\_phrases}
    \item 30\% \textit{desire\_phrases}
    \item 20\% \textit{problem\_phrases}
    \item 20\% \textit{other patterns}
\end{itemize}

Other aspects of the data are the same as described in \ref{data_an}.

% \begin{enumerate}
%     \item Upsample purpose of call: 42.5\% negative, 42.5 positive, 15\% questions\footnote{roughly correspond to the proportion of call purpose utterances following a question in real data}, 100\% utterances from calls from hits
%     \item Upsample purpose of call: 42.5\% negative, 42.5 positive, 15\% questions; include 25\% calls without hits\footnote{the approximate rate of “negative” hits based on hitrate}
%     \item Balanced classes: 33.3\% each class, including 25\% calls without hits.
%     \item Random sample: 4.7\% positive, 93.4\% negative, 1.9\% questions, 95\% utterances come from calls without hits
%     \item Upsample purpose of call but keep the prevalence of the negative class: 20\% call purpose, 65\% negative, 15\% questions, 25\% calls without hits.
% \end{enumerate}

\section{Appendix: Training details}
\label{appendix_training}

\textbf{Parameters:}
The pretrained \textit{distilbert-base-cased} model we use has 6 layers, 768 hidden units, 12 attention heads and 65M parameters and is available through Multimodal-Toolkit~\cite{gu-budhkar-2021-package}.
We run all fine-tuning experiments on a Google Cloud VM n1-standard-8 instance with 496GB disk size and 1 NVIDIA Tesla K80 GPU. The maximum time for a single experiment was 8 GPU hours.
We truncate text input to a maximum 150 tokens since most relevant statements fall into this category. We set the train and eval batch size to 32 and 64 respectively, and use AdamW optimizer with default parameters. We fine-tune the model for 4 epochs with a learning rate of 5e-05, weight decay of 0.01 and 500 warm up steps. The hyperparameters were obtained from experiments using an in-house tuning tool implementing grid search algorithm. For fine-tuning on small subsets (4K) of data collected through error analysis, we repeat the training process for 12 epochs and a learning rate of 9e-05.

\textbf{Relevant features:}
Besides the text features, we experimented with two extra features that have proven to be useful in the knowledge-engineered model: the \textit{start time} of the utterance and the \textit{call side}. We also experimented with several mechanisms to combine the numerical and categorical features with textual data using Multimodal-Toolkit~\cite{gu-budhkar-2021-package}. The results are presented in Table \ref{features}.

\begin{table}[hbt!]
\centering
\begin{tabular}{p{0.33\linewidth} | p{0.1\linewidth} | p{0.1\linewidth} | p{0.1\linewidth} | p{0.1\linewidth}}
\hline
\textbf{Feature} & \textbf{P} & \textbf{HR} & \textbf{F1} & \textbf{PP}\\
\hline
text only & 0.891 & 0.891 & 0.891 & 0.894 \\
text + start time & 0.948 & 0.948 & 0.948 & 0.944 \\
text + call side & 0.948 & 0.948 & 0.948 & 0.946 \\
all & 0.949 & 0.949 & 0.949 & 0.957 \\
\hline
\end{tabular}
\caption{\label{features}
Comparing model performance using tabular features \textit{start time} and \textit{call side}. P-Precision, HR-Hit rate, PP - precision in \textit{positive} class. The results are reported for a single run using concatenation to combine features.
}
\end{table}

\end{document}